\documentclass{article}
\usepackage{ijcai21}

\usepackage{times}

\usepackage{soul}
\usepackage{url}
\usepackage[hidelinks]{hyperref}
\usepackage[utf8]{inputenc}
\usepackage[small]{caption}
\usepackage{graphicx}
\usepackage{amsmath}
\usepackage{booktabs}
\usepackage{xcolor}

\title{VisioRed: A Visualisation Tool for Interpretable Predictive Maintenance}

\author{
Spyridon Paraschos\footnote{Contact Author}\and
Ioannis Mollas\and
Nick Bassiliades\And
Grigorios Tsoumakas\\
\affiliations
Dept. of Informatics, Aristotle University of Thessaloniki, Greece\\
\emails
\{pspyrido, iamollas, nbassili, greg\}@csd.auth.gr
}

\begin{document}

\maketitle

\begin{abstract}
The use of machine learning rapidly increases in high-risk scenarios where decisions are required, for example in healthcare or industrial monitoring equipment. In crucial situations, a model that can offer meaningful explanations of its decision-making is essential. In industrial facilities, the equipment's well-timed maintenance is vital to ensure continuous operation to prevent money loss. Using machine learning, predictive and prescriptive maintenance attempt to anticipate and prevent eventual system failures. This paper introduces a visualisation tool incorporating interpretations to display information derived from predictive maintenance models, trained on time-series data.
\end{abstract}

\section{Introduction}

The majority of research conducted to explain machine learning (ML) models focuses on the tasks of image recognition and object identification using feature maps of the convolutional layers in a deep neural network to capture its visual attention~\cite{samek2019explainable}. In contrast, only a few works have researched the explainability of models trained on time series (TS)~\cite{karlsson2018explainable}.

There exist numerous visualisation methods for image related tasks~\cite{yuan2020survey}. On the other hand, visualisation methods for TS forecasting models are scarce and exhibit limitations. Visplause~\cite{arbesser2016visplause} for example, lacks support for local interpretation and has a complex user interface. The visualisation tool in~\cite{assaf2019explainable} is based on convolutional neural networks and gradient-based interpretation techniques for TS models. These techniques, however, are specifically designed for networks containing only 1D convolutional layers and cannot be generalised to other architectures, while their applicability is limited to classification problems.

\begin{figure}[ht]
\centerline{\includegraphics[width=0.5\textwidth]{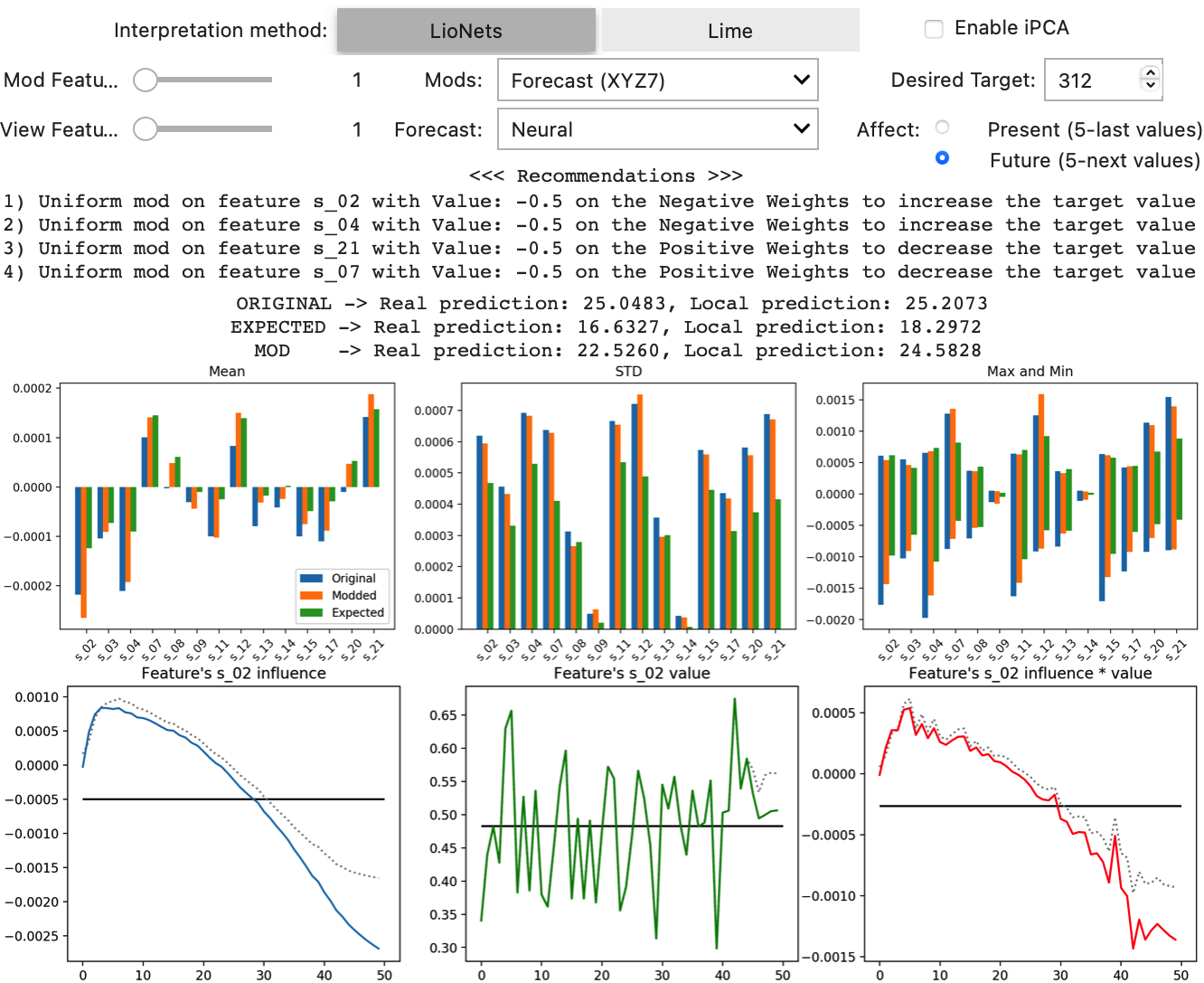}}
\caption{VisioRed template and XYZ example}
\label{fig:visiored_tool}
\end{figure}

This paper presents VisioRed, a system for visualising interpretations of TS model predictions, which includes two main novel techniques~\cite{SpyrThesis}. The first one, iPCA, concerns the inclusion of a dimensionality reduction technique in its pipeline that gives valuable latent information on the value of its features. The second one, called XYZ, concerns a model for conditional forecasting of future feature values given past values and, crucially, a preferred target value. 

We focus specifically on predictive maintenance (PdM), a very important and popular task nowadays in the industrial sector. PdM systems can predict with high precision when repairs or replacements to equipment are required, reducing additional costs by preventing unnecessary repairs~\cite{predictive_maintenance}. In most PdM tasks, the objective is to predict a component's remaining useful lifetime (RUL), thus we consider regression problems. Typically, PdM involves the analysis of various TS produced by multiple monitoring sensors. Hence, an explanation that includes not only the sensor information, to identify the source of the problem, but also temporal information should often be provided to clarify the importance of measurements at each time step. Prescriptive maintenance (PsM), exploiting such information, evolves the PdM concept by providing suggestions about how to prevent early failure through maintenance or other actions, along with the predictions~\cite{khoshafian2015digital}.

\section{VisioRed}
\label{ch:section_4}
We propose VisioRed as a visualisation tool for assisting the user to understand a model's prediction, by incorporating a set of modifications, recommendations, forecasters, and interpretability options, as well as some additional functionalities towards PsM. All the interpretability methods, the black box predictive model (PM) and the experimentation tools that we use and develop are compatible with multi-variate TS data, for the final product to be applicable regardless of their nature. Therefore, the supported dimension of an input instance $x_i$ is a $J\times N$ matrix, where $J$ is the number of features of the TS and $N$ is the number of consecutive selected time steps, while the output is the RUL prediction.

\begin{equation}
x_{i} = 
\begin{pmatrix}
v_{0,0} & v_{0,1} & \cdots & v_{0,N-1} \\
v_{1,0} & v_{1,1} & \cdots & v_{1,N-1} \\
\vdots  & \vdots  & \ddots & \vdots  \\
v_{J-1,0} & v_{J-1,1} & \cdots & v_{J-1,N-1} 
\end{pmatrix}
\end{equation}

\subsection{Local Explanation with Feature Importance}

VisioRed includes two local explanation techniques, LioNets~\cite{lionets} and LIME~\cite{lime}. Given a new sequence instance, $x_i$, these techniques generate a set of neighbours $L(x_i)$ and then build a transparent linear model to extract feature importance for explaining the prediction of the PM. Both techniques assign $J\times N$ importance values $ts_{j,n}$, one for each time step of each feature. However, we assume that users would rather avoid this complexity and examine the importance $s_j$ of each feature. A straightforward way to achieve this, is to present users $s_j=\frac{1}{N} \sum_{n=1}^{N} ts_{j,n}$; the mean of the importance of all the time steps of each feature. 

A first innovation of VisioRed is an alternative approach to summarising the importance of a feature, based on principal components analysis (PCA), called iPCA (interpretations through PCA)~\cite{SpyrThesis}. Given a new sequence instance, $x_i$, iPCA applies PCA once for each feature, on the data comprising the time steps of that feature in $x_i$ and its neighbours $L(x_i)$, which are created by the local explanation technique (LioNets or LIME). By taking the first principal component of these PCA transformations, we obtain a distilled representation with dimensionality $J$ instead of $J \times N$ for both $x_i$ and its neighbours $L(x_i)$. Then, the local explanation technique trains the linear model on these reduced representations. The feature importance extracted from this transparent model will correspond directly to the features, bypassing the need for aggregation. 


Experimental results using a variety of metrics suggest that iPCA captures the importance of features better or equally well than averaging the individual importances of all time steps~\cite{SpyrThesis}. We believe that this happens because the distilled representation of the input instances makes the approximation task of the linear model easier. Moreover, iPCA can still assign an importance value to the time steps of each feature, based on the coefficients of the respective PCA feature transformation. These coefficients do not correspond to actual time step weights, but they provide a measure of the contribution of each time step to the formation of the latent representation of a feature, and by extension of the influence of each time step to the prediction. 


\subsection{Modification Recommendations}
Towards PsM, VisioRed allows users to experiment with the values of an instance, in order to test what-if scenarios and explore their influence on the model's decisions. Given a new sequence instance, $x_i$, {\em modifications} concern changing the values in a sub-sequence of a feature's time steps. The available modification types are: a) adding uniform noise, b) adding Gaussian noise, c) replacing with mean value, and d) replacing with zeros. Users select a feature, a sub-sequence of time steps and a modification type and view the influence of the changes to the RUL.

Due to the very large number of different modifications that users can explore, VisioRed offers a collection of four modification {\em recommendations}. Each recommendation involves one feature and one modification. The features involved in these recommendations are the two features with the highest negative importance and the two features with the highest positive importance. 
For each of these features, VisioRed automatically tests all four modifications and chooses those that are expected to increase or decrease the RUL most.

\subsection{Conditional Forecasting of Feature Values}

\begin{figure}[ht]
\centerline{\includegraphics[width=0.5\textwidth]{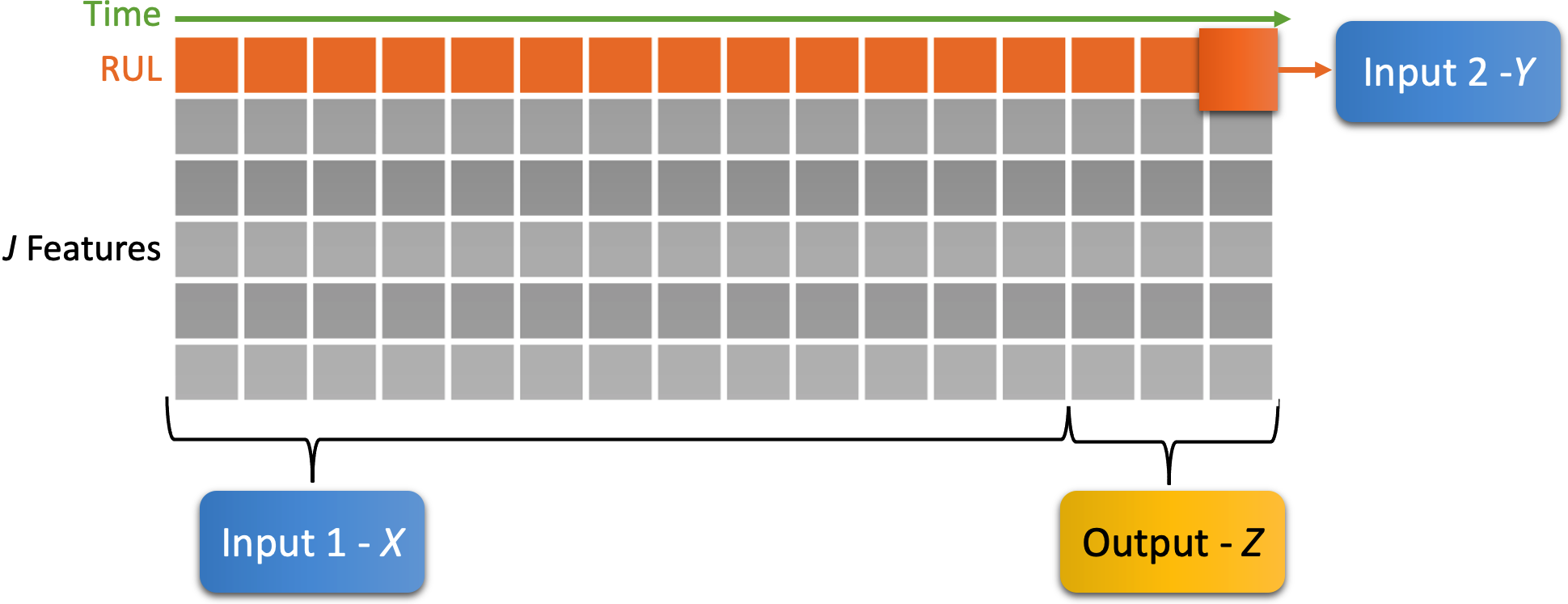}}
\caption{XYZ inputs and output}
\label{fig:xyz}
\end{figure}

Modification recommendations allow users to explore the relationship of individual features with the target. To take this PsM approach a step further, VisioRed includes an additional neural model, called XYZ, which allows users to enter a higher RUL than the currently predicted one, and outputs the values that the features should have in the next time steps. XYZ is trained using as input the first $X$ time steps of the features of an instance and the prediction of the PM for that instance $Y$, and as an output the last $Z$ time steps of that instance, as shown in Figure~\ref{fig:xyz}. During inference, users can provide the first $X$ values of an instance as well as the preferred target $Y$ value, and the model will suggest how the $Z$ time step measurements should be in order to reach the preferred $Y$ value. 

In addition, VisioRed provides three forecasting options to allow users to explore the potential $Z$ future time step values of the features and their effect to the future prediction, as well as to complement the functionality of XYZ. These are a custom neural forecaster (NF), the N-Beats~\cite{oreshkinCCB20} forecaster and a static forecaster (SF). The neural architecture of NF is similar to that of the PM, as this leads to high performance as validated through our experiments. As for the SF, the idea is to use a fast and simple model as a baseline. The SF is a non-neural solution, that just copies the last $Z$ time steps and rotates them by 180 degrees. 

Combining XYZ with the forecasters, the user gets the best of both worlds. Providing as input the $X$ latest time step measurements and a preferred $Y$ value, the user can compare the output of XYZ with the output of a forecaster, in order to see which feature measurements require intervention in order to achieve the requested $Y$ value.

\subsection{User Interface}
Figure~\ref{fig:visiored_tool} shows the user interface (UI) of VisioRed that integrates all the methods discussed earlier in this section. By selecting the preferred interpretation technique the exploration of an instance's prediction is initiated. Then, the UI allows the user to observe individual feature statistics, derived from the interpretation, and provide all the functions required to alter the application window and its measurements. Each modification setup can require its own parameters, so further necessary controls appear when needed. 

A separate area contains the PM's prediction of an instance, the interpretation's local prediction, and predictions for the modifications applied to the instance. Furthermore, two sets of plots are available. The first set shows the cumulative feature importance of the features, while the second highlights the importance per time step for a selected feature. Using a dedicated view slider, the user can display the time steps of any feature. Moreover, these plots include both the original interpretation of the examined instance as well as the interpretation obtained after a modification.

\section{Experiments}
\label{ch:experiments}

Having defined the core ingredients of VisioRed, we will now evaluate its usefulness using the turbofan engine degradation dataset~\cite{TEDS,RULotherPaper}, which contains a multi-variate TS for PdM. The dataset initially contained measurements for 32 sensors, which after feature selection were reduced to 14. The target value is the RUL of the engine, which determines how long the engine can run properly before a malfunction occurs. The code of the following examples is available in VisioRed's GitHub\footnote{\url{https://git.io/JqLci}} and DockerHub\footnote{\url{https://dockr.ly/3cJ4Qx0}} repository.

Therefore, by utilising a PM and given the measurements of the sensors within a certain time-window, we want to predict the RUL of an engine, in order to prevent an imminent failure. Before proceeding with the training of our models, we define the input dimension as a $50\times14$ matrix, where $50$ is the number of time steps (measurements) and $14$ the number of features (sensors) and then we extracted individual instances, from both the training and testing set, based on these dimensions. The target value of PM is the value of RUL at the final time step of each instance.

Using a trained PM, we train the forecaster and XYZ models with $Z=5$ forecast window. We also tune the interpretation techniques, LioNets and LIME, with and without iPCA, using the fidelity and truthfulness~\cite{altruist} interpretability metrics. The models used in these experiments are not extensively trained, as our goal was not to achieve top notch performance, but rather to showcase a variety of interesting tool ideas to experiment with the time step values.

\subsection{Prescriptive Maintenance Example}
We will now show an example of experimenting with VisioRed's XYZ model towards prescriptive maintenance of a turbofan engine. Right after loading a random instance we can see how the tool appears to the user in Figure~\ref{fig:visiored_tool}. The RUL that the model predicted is $25.05$ time steps, while LioNets predicted a similar value of $25.21$. The predicted RUL is alarming, since it suggests that the engine will malfunction in a small number ($25$) of time steps.

Given a forecast window of $5$ steps and setting a desired target of $312$ time steps to prolong the engine's lifetime, XYZ informs us what the sensors' values should be. Having that information, we can make immediate temporary repairs to the engine, so that the sensor readings range as close as possible to the ones that the XYZ model suggests. In the tool, we select the ``Forecast (XYZ)'' and the ``Future'' option, because we want XYZ to generate a prediction for the upcoming $5$ time steps. As a forecaster, we keep the default option, the NF. We define the ``Desired Target'' to $312$. We most definitely want to optimise the RUL, so we set the target to the highest possible RUL value. In Figure~\ref{fig:visiored_tool}, the results of the XYZ are shown. After running XYZ, three separate predictions appear. The first is the original RUL prediction, while the second concerns the RUL value after $5$ time steps, as forecasted by the NF ($16.63$). The third prediction, ``MOD'', is the RUL value that we would get if the observations of the sensors are in agreement with what XYZ has proposed.

In our example, it is evident that the third prediction ($22.53$) is not comparable to the desired target we defined ($312$), and anyone could draw the misconclusion that the XYZ model did not succeed. However, even in that case, XYZ managed to provide a set of measurements which will increase the expected RUL from $16.63$ to $22.53$. We, therefore, believe that this outcome would help in real case scenarios to prepare more realistic fixes to the targeted machinery, given the fact that an additional breathing space of approximately 6 time steps would be added to its RUL.

\section{Conclusions}
In this paper, we introduced interpretability to the PdM pipeline via the implementation of novel interpretation methods and a new TS specific approach, named iPCA. A visualisation tool, called VisioRed, with capabilities including modifications, recommendations and forecasting was developed to elevate the interpretability of PdM towards PsM. An experimental study of VisioRed, demonstrating its efficacy and effectiveness, has been carried out with a well-known PdM dataset. In the future, we seek to explore the applicability of VisioRed in non-regression ML tasks, such as binary or multi-class classification.

\bibliographystyle{named}

\end{document}